\documentclass{article} % For LaTeX2e
\usepackage{iclr2026_conference,times}

\usepackage{amsmath,amsfonts,bm}

\def\eqref#1{equation~\ref{#1}}
\def\1{\bm{1}}

\DeclareMathAlphabet{\mathsfit}{\encodingdefault}{\sfdefault}{m}{sl}
\SetMathAlphabet{\mathsfit}{bold}{\encodingdefault}{\sfdefault}{bx}{n}

\usepackage{hyperref}
\usepackage{url}
\usepackage{multirow}
\usepackage{booktabs}
\usepackage{xcolor}
\usepackage{tabularx}
\usepackage{graphicx}
\usepackage{tcolorbox}

\title{KBE-DME: Dynamic Multimodal Evaluation via Knowledge Enhanced Benchmark Evolution}
\author{Junzhe Zhang, Huixuan Zhang \& Xiaojun Wan \\
Wangxuan Institute of Computer Technology\\
Peking University\\
\texttt{\{junzhezhang, zhanghuixuan\}@stu.pku.edu.cn}\\
\texttt{\{wanxiaojun\}@pku.edu.cn}
}
\iclrfinalcopy 
\begin{document}

\maketitle

\begin{abstract}
The rapid progress of multimodal large language models (MLLMs) calls for more reliable evaluation protocols. Existing static benchmarks suffer from the potential risk of data contamination and saturation, leading to inflated or misleading performance evaluations. To address these issues, we first apply Graph formulation to represent a static or dynamic VQA sample. With the formulation, we propose Knowledge-enhanced Benchmark Evolution(KBE), a dynamic multimodal evaluation framework. KBE first analyzes the original static benchmark, then expands it by integrating multimodal knowledge, transforming the static benchmark into a controllable, dynamic evolving version. Crucially, KBE can both reconstruct questions by Re-selecting visual information in the original image and expand existing questions with external textual knowledge. It enables difficulty-controllable evaluation by adjusting the degree of question exploration. Extensive experiments demonstrate that KBE alleviates the risk of data contamination, data saturation, and provides a more comprehensive assessment of MLLM capabilities.
\end{abstract}

\section{Introduction}
\label{sec:intro}
%思路
%从数据污染和数据饱和入手?
%现存的静态评测基准承受着数据饱和和数据污染的风险

%多模态大模型近期发展迅速，这些模型在许多任务上都表现良好。由此，越来越多的测试基准被提出用来评测不同多模态大模型的能力。其中有一些静态数据集设计的全面和solid，一定程度上反映了现有的多模态模型在不同多模态任务上的表现能力。
Multimodal large language models(MLLMs) have been developing rapidly in recent years, demonstrating remarkable performance across a wide range of tasks\citep{li2024llavaonevisioneasyvisualtask, bai2025qwen25vltechnicalreport}. This rapid progress has motivated the creation of an increasing number of multimodal benchmarks designed to evaluate their capabilities. Several traditional static benchmarks\citep{marino2019ok, schwenk2022aokvqabenchmarkvisualquestion} have been carefully curated with comprehensive testing coverage and rigorous construction processes. These benchmarks provide evidence of how current multimodal models perform across diverse multimodal tasks, and are crucial in understanding the strengths and weaknesses of MLLMs.

%然而，目前的评测方式也存在一些隐含的问题。第一点是潜在的数据污染问题。上述提到的开源测试集中，大部分都会提供自己的测试数据用来测试不同的多模态大模型。这样做保障了测试的方便性和可复现性，然而也增加了测试数据泄露的风险。随着静态数据集的测试集被开源，即使模型不去有意的获取测试数据，这些测试数据也可能流入到模型的训练数据中。由于数据污染的风险，静态开源benchmark的评估有效性会随着开源时间的增加而受到影响。
However, current evaluation practices also suffer from several implicit limitations. A primary concern lies in the risk of data contamination\citep{song2025textimagesleakedsystematic}. Most of the aforementioned open-source benchmarks release their test samples and labels to facilitate reproducibility of comparison across different multimodal large models. While this openness ensures transparency, it also increases the risk that open-sourced test data may be inadvertently leaked into training corpora of existing MLLMs. The wide availability of benchmark test sets means that portions of these datasets can be unintentionally included during large-scale pretraining. As a result, the reliability of static open-source benchmarks gradually diminishes over time, since their effectiveness as unbiased evaluators is compromised by potential overlap with training data.
%OK-VQA\citep{marino2019ok} and A-OKVQA\citep{schwenk2022aokvqabenchmarkvisualquestion}
%找数据污染的一些citation
%另外一个潜在问题是数据饱和，目前的多模态大模型本身的能力进步很快，模型的表现在一直增加，而静态benchmark的题目难度无法随之一起进化，导致部分模型在一些数据集上取得了较好的成绩\citep{MMBench}，使得整体的题目难度无法匹配性能提升后的模型。此时，原先固定难度的benchmark对于新模型的区分度就减弱了。
Another issue lies in data saturation. As multimodal large language models continue to develop rapidly, their performance on many established benchmarks keeps improving. However, the difficulty static benchmarks remains fixed and cannot evolve along with the increasing capabilities of newer MLLMs. As a result, certain models have already achieved high scores on some widely used datasets OK-VQA\citep{marino2019ok}, raising concerns about the diminishing discriminative power of such benchmarks. In this scenario, benchmarks that were once sufficiently challenging can no longer provide a reliable separation between state-of-the-art systems, thereby limiting their utility in driving future progress.

%为了解决这两个问题，传统的做法是根据目前的MLLMs的性能及时的创造特定的难度的benchmark，来暂时进行难度控制和规避数据污染风险。然而，如之前所说，这样构建的benchmark的有效性会随着时间的推移而减弱，而反复的构建特定难度的Benchmark会耗费大量的人力和时间。只有使用随着MLLMs一起进化的Benchmark对MLLMs进行动态评估才能解决上述传统静态评价体系中的两个问题。
%反复的构建难度更高的静态基准比较费时间 
To address these concerns, a conventional approach is to constantly design new benchmarks whose difficulty is suitable for evaluating the current capabilities of MLLMs. Such new constructed benchmarks can temporarily control task difficulty and mitigate the risks of data contamination. However, the validity of these benchmarks inevitably diminishes over time, and the repeated construction of specialized static datasets requires substantial human effort and is time-consuming. We argue that a more sustainable solution is to develop frameworks that enables benchmarks to evolve with MLLMs, enabling dynamic evaluation. Only through such adaptive benchmarks can we overcome the dual limitations of data contamination and data saturation inherent in the traditional evaluation with static benchmark.

\begin{figure}[t]
\begin{center}
%\framebox[4.0in]{$\;$}
    \includegraphics[width=0.9\linewidth]{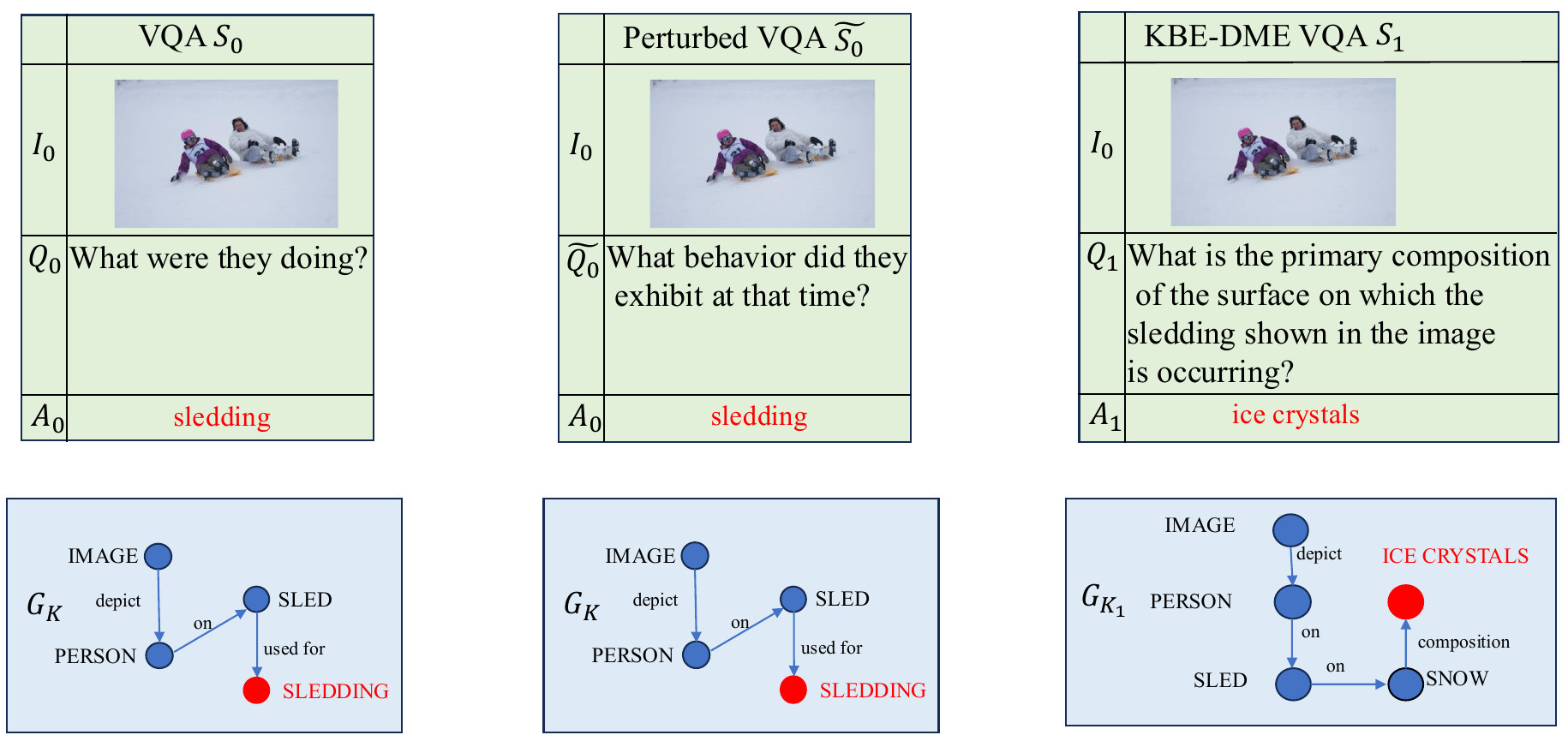}
    \caption{Figure of the original VQA test sample $S_0$, the perturbed test sample $\tilde{S_0}$ generated by perturbation methods, and the dynamically generated test sample $S_1$ produced by KBE-DME. Here, $G_K$ denotes the key information subgraph extracted from a VQA test sample. As shown, the perturbed VQA test sample does not alter the underlying $G_K$ of the question.}
    \label{fig:intro1}
\end{center}
\end{figure}

%这个图展示了原始的VQA测试数据S_0，以及基于扰动方法生成的VQA测试数据\hat{S_0}和使用KBE-DME动态生成的VQA测试数据S_1。其中G_K是我们对于一个VQA测试数据提取的关键信息子图。可以看到，扰动生成的VQA测试数据并没有改变问题的G_K。

%为了解决数据污染和数据饱和问题，我们提出了KBE-DME: DYNAMIC MULTIMODAL EVALUATION VIA KNOWLEDGE ENHANCED BENCHMARK EVOLUTION。我们首先用VQA格式表示多模态任务的问题，然后使用图和三元组的定义来建模一个VQA问题。我们会生成一个VQA问题所包含的潜在相关的多模态知识三元组，并从中选择需要回答这个VQA问题所需要的关键三元组。之后，我们会通过1.从潜在知识三元组中重新选择关键三元组，2.通过外部知识增加新的知识三元组到关键三元组，两种方式来改变关键三元组。而后，KBE-DME结合经过改变的关键三元组与原始问题的VQA信息综合生成新的VQA问题用来进行动态评估。

There are some dynamic evaluation methods\citep{yang2025dynamicmultimodalevaluationflexible} based on perturbation. However, the perturbed VQA test sample sometimes does not alter the core of the question as shown in Figure \ref{fig:intro1}. To overcome the challenges of data contamination and data saturation, we introduce KBE-DME: Dynamic Multimodal Evaluation via Knowledge-Enhanced Benchmark Evolution. Our approach starts with the multimodal tasks represented in the standard VQA format, and further modeling each VQA problem using a Graph formulation with mutlimodal knowledge triplets. For every question, we extract a set of candidate multimodal knowledge triplets and then identify the subset composed of key triplets that are necessary for answering the question. Based on this representation, KBE-DME introduces a framework dynamically evolves benchmarks in two different ways: 1.Re-selection of key triplets from the pool of candidate multimodal knowledge triplets, which indicates the reasoning rationale required to answer the question; 2.Exploration with external knowledge triplets, where new triplets are incorporated into the key triplets set to enrich the knowledge required.
By integrating the modified key triplets with the information of original VQA sample, KBE-DME synthesizes novel, knowledge-enhanced VQA questions with controllable difficulty. These dynamic questions can continuously evolve with the progress of multimodal large models through our framework. This dynamic construction not only mitigates the risks of test set leakage but also ensures that the difficulty of evaluation adapts to the improving capabilities of MLLMs.

%experiments说我们的方法的泛化性和进行的实验
%我们的方法KBE-DME具有良好的泛化性，能够对不同的多模态VQA数据集进行改造从而实现动态评估。我们在OK-VQA和AOK-VQA这两个VQA数据集上测试了KBE-DME，并测试了5个不同的多模态大模型。我们进行了充分的模型测试和生成数据的分析实验。结果证明KBE-DME可以基于原始静态测试数据动态的构建不同难度的测试数据，从而对待测试的多模态大模型进行难度可控的动态评估。
Our proposed method, KBE-DME, exhibits strong generalization ability and can be applied to transform different multimodal VQA datasets for dynamic evaluation. We apply KBE-DME on two widely used benchmarks, OK-VQA\citep{marino2019ok} and A-OKVQA\citep{schwenk2022aokvqabenchmarkvisualquestion}, and evaluate five representative MLLMs using our dynamic evaluation framework. Through extensive experiments and analyses of the generated dynamic test data, our results demonstrate that KBE-DME is capable of dynamically constructing test sets of various difficulty levels based on static benchmarks. This enables a dynamic and difficulty-controllable evaluation of MLLMs, effectively overcoming the limitations of traditional static benchmarks.

Our contributions are:
\begin{itemize}
    \item We introduce a novel graph formulation to represent a VQA sample, where the multimodal knowledge is represented as triplets. Within this formulation, the process of dynamic test data generation for evaluation can be naturally expressed as transformations of triples in the graph structure. 
    \item We propose KBE-DME, a dynamic evaluation framework which evolves static multimodal benchmarks via re-selection and exploration strategy, generating difficulty-controllable test data that co-evolves with the progress of MLLMs.
    \item We conduct extensive experiments with KBE-DME on two static VQA benchmarks, evaluate five representative MLLMs and perform detailed analyses of the dynamically generated data. The results demonstrate that KBE-DME enables high-quality and difficulty-controllable dynamic evaluation of MLLMs.
\end{itemize}

%我们提出使用Graph Formulation的形式来抽象一个VQA问题，并将动态评估中动态测试数据的生成表示为原有问题在图结构上三元组的变化。
%我们提出了KBE-DME动态评价framework，他可以解析不同静态多模态评价数据，并通过Re-selection或者是Exploration基于静态的评价数据来动态生成难度可控的评测数据，从而使得原先静态的评测数据可以随着模型的发展而一同进化。
%我们使用KBE-DME在两个静态VQA数据上进行了充分的实验，包括测试5个主流的多模态大模型的表现并对动态生成的数据进行了细致的分析。实验和分析结果表明，KBE-DME能够基于静态VQABenchmark对待测试的MLLMs，进行高质量和难度可控的动态评估。

%contribution

%exploration experiments mutli-hop experiments
%main performance 
%[veiry model] [ablation study] [GPT-直接生成(结果不好分析)]
%Gemini 测试
%quality check [verify model]
%data distribution 

\section{Related Work}
\label{sec:related_works}
\subsection{Static Multimodal Evaluation}
%静态的多模态评估作为一种传统的评价手段一直再发展。
%随着近些年来多模态大模型的进步，为了测试新的高性能模型，越来越多的静态评测数据集被提出。
Static multimodal evaluation has long served as the standard paradigm for assessing the capabilities of multimodal models. Early efforts introduced benchmarks such as MSCOCO Captions\citep{chen2015microsoftcococaptionsdata}, VQA-v2\citep{Goyal_2017_CVPR}, OK-VQA\citep{marino2019ok}, TextVQA \citep{textvqa}, DocVQA \citep{docvqa}, InfoVQA \citep{infovqa} and A-OKVQA\citep{schwenk2022aokvqabenchmarkvisualquestion}, which provided fixed datasets and standardized metrics to measure abilities such as visual understanding, reasoning, and image–text alignment. These static benchmarks played a crucial role in model developing progress by offering a common ground for model comparison.
With the rapid advancement of multimodal large models (MLLMs) in recent years\citep{li2024llavaonevisioneasyvisualtask, bai2025qwen25vltechnicalreport}, previous proposed benchmarks are no longer sufficient to meet the need for evaluating increasingly powerful MLLMs.
To keep pace with increasingly powerful models, plenty of new static benchmarks have been proposed, aiming to provide broader coverage and more challenging tasks. For example, some specific benchmarks such as ChartQA\citep{masry-etal-2022-chartqa} focuses on chart understanding; and others such as MMBench\citep{liu2024mmbenchmultimodalmodelallaround}, MME\citep{fu2024mmecomprehensiveevaluationbenchmark},  MMStar \citep{mmstar} and SEED-Bench\citep{li2023seedbenchbenchmarkingmultimodalllms} provide comprehensive multi-dimensional evaluations like reasoning, OCR, and others.

Despite their success, static multimodal benchmarks remain inherently constrained by fixed difficulty and potential data contamination once released\citep{yang2025dynamicmultimodalevaluationflexible}.
Although some studies have attempted to change their evaluation questions\citep{shah2019cycle, 10.1007/978-3-030-58589-1_23}, these methods are typically designed for specific datasets and are difficult to serve as a widely applicable dynamic evaluation strategy for other multimodal static benchmarks.
As models continue to evolve, even carefully curated datasets may gradually lose their discriminative power, highlighting the need for dynamic and adaptive evaluation paradigms that can co-evolve with model capabilities.

%%尽管一些工作尝试动态的变换他们的问题，但是他们的方法往往是针对自己的数据集，并不能作为一种动态评估策略广泛的推广到所有的数据集上。

% OK-VQA and A-OKVQA extend evaluation beyond perception to incorporate external knowledge; NLVR2 focuses on logical reasoning across paired images; and more recent efforts such as MMBench, MME, and SEED-Bench provide comprehensive multi-dimensional evaluations, spanning perception, reasoning, OCR, and dialogue. These datasets reflect the community’s attempt to design more rigorous and holistic evaluation standards for the growing diversity of MLLM capabilities.

%\section{Problem Formulation}
%Problem Formulation
\subsection{Dynamic Evaluation}
To mitigate the data contamination and data  saturation issues, recent studies have explored dynamic evaluation\citep{jiang2025raising, yang2025dynamicmultimodalevaluationflexible}, where test data are perturbed\citep{yang2025dynamicmultimodalevaluationflexible} or regenerated\citep{jiang2025raising} to adapt difficulty and reduce data contamination effects. In the field of text-only dynamic evaluation, DyVal\citep{zhu2024dyvaldynamicevaluationlarge} dynamically generate test samples to mitigate data comtamination. NPHardEval\citep{fan-etal-2024-nphardeval} generate new samples for NP-hard math problems evaluation. MPA\citep{zhu2024dynamicevaluationlargelanguage} apply agent to generate new evluation samples. 

However, in the multimodal domain, research on dynamic evaluation remains relatively limited. VLB\citep{yang2025dynamicmultimodalevaluationflexible} represents one of the first attempts to bootstrap both images and text simultaneously by editing objects or backgrounds in images, replacing or rephrasing words in questions, and adding related or unrelated textual content to perturb the original VQA problems. \citet{liu2025reasoningmultimodallargelanguage} proposes a multimodal dynamic evaluation framework to perturb the multimodal task itself instead of perturbing inputs. 
While perturbation-based methods indeed modify the test inputs, their impact is relatively limited compared to regenerating entirely new test data. As a result, the scope of dynamically generated data remains constrained. Moreover, perturbation approaches offer little control over the difficulty level of the generated data. To achieve both a broader range of dynamically generated data and finer-grained difficulty control, we propose KBE-DME, a dynamic multimodal evaluation framework that regenerates new test data instead of merely perturbing existing ones.

%然而，在多模态领域，相关的动态评估工作还不多。VLB首次对图像和文本同时进行bootstrap，也就是对图像物体背景的编辑，和替换词语，rephrase问题，增加相关或无关的文本来对原始VQA问题进行扰动。

%扰动确实改变了测试问题的输入，然而比起重新生成新测试数据，对原始数据的扰动所造成的改变较为有限，是得动态生成的数据范围受到限制。另外，扰动类型的方法也不好进行难度控制，因此为了获得更大的动态生成数据的变化范围，以及对动态生成的数据进行更精细的难度控制，我们提出了基于Graph Formulation的KBE-DME方法来重新生成新的测试数据，而不是只对原有数据进行扰动。

%在多模态领域，除了扰动相关的方法，

%While these approaches demonstrate the potential of dynamic benchmarks to co-evolve with multimodal large models, most remain limited in scale or scope, leaving room for more systematic frameworks.

%一些尝试
%单模态

%多模态

\begin{figure}[t]
\begin{center}
%\framebox[4.0in]{$\;$}
    \includegraphics[width=0.9\linewidth]{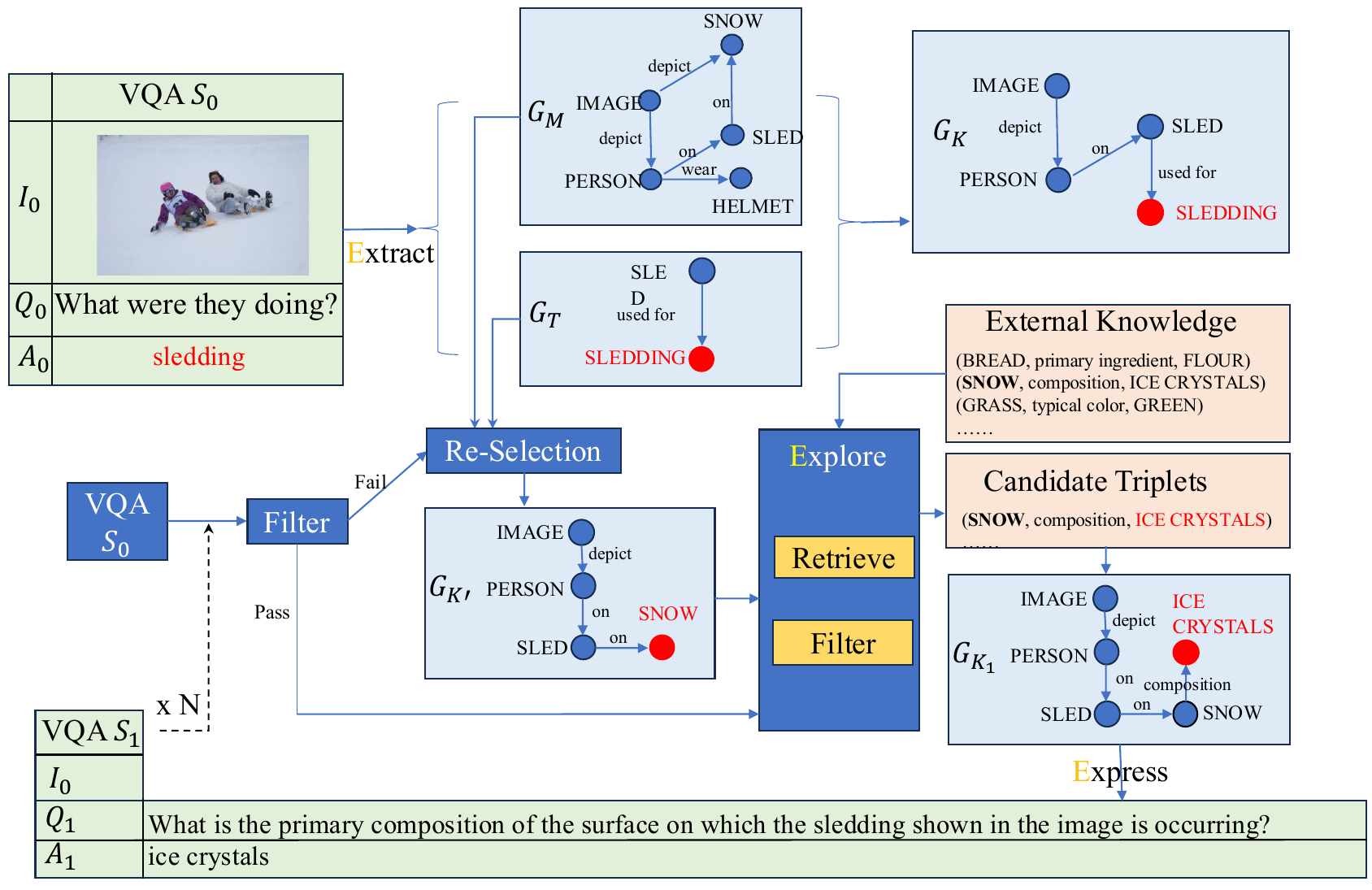}
    \caption{Figure of our Graph Formulation and the KBE-DME framework. The upper part of the figure uses a static VQA sample $S_0$
  to exemplify our graph representation of a VQA problem, while the lower part demonstrates the KBE-DME framework for dynamically constructing VQA test data.}
    \label{fig:method}
\end{center}
\end{figure}

%这是展示我们的Graph Formulation和KBE-DME framework的图。图的上半部分以一个静态VQA数据S_0举例说明了我们对于一个VQA问题的图表示，图的下半部分则说明了我们的KBE-DME的动态构造VQA测试数据的框架。

\section{KBE-DME}
\label{sec:KBE-DME}
\subsection{Graph Formulation}
%我们使用Graph的Formulation来表示一个VQA形式的多模态问题。一个问题的Graph Formulation由不同的多模态知识三元组组成。一条知识可以以多模态知识三元组(s, r, o)表示。其中s是三元组triplet的主体，r是对应的关系，o则是主体s在对应关系r下的对应结果。

We represent a multimodal VQA problem using a graph formulation, where each problem is abstracted as a structured graph composed of multiple multimodal knowledge triplets. 

\paragraph{Knowledge Triplet} A unit of knowledge can be represented as a multimodal knowledge triplet $(s,r,o)$. $(s,r,o)$, where $s$ denotes the subject of the triple, $r$ specifies the relation, and $o$ corresponds to the object associated with $s$ under relation $r$. 

%我们将知识三元祖中的s或o作为图中的节点，将对应的三元组(s, r, o)作为边分别构建一个VQA问题的Visual Graph G_V和Textual Graph G_T。
\paragraph{Graph Definition of Static VQA}
We treat the $s$ and $o$ in a knowledge triplet as vertices in the graph, and use the corresponding triplet $(s, r, o)$ as an directed edge connecting $s$ and $o$. We first construct visual knowledge triplets $M = \{(s_m, r_m, o_m)\}$ to represent the visual information of a VQA sample, and then construct textual knowledge triplets $T = \{(s_t, r_t, o_t)\}$ to represent the worldwide textual background knowledge of the VQA sample. Based on this representation, we construct a Visual Graph $G_M = <V_M, E_M>$ with visual knowledge triplets,
which is:

\begin{equation}
    V_M =\{s,o \mid \exists (s, r, o)\in M\}, E_M =\{(s_{m}, o_{m},[r_{m}]) \mid \exists (s_{m}, r_{m}, o_{m})\in M\}
\label{eq:E_M}
\end{equation}

and a Textual Graph $ G_T  = <V_T, E_T>$ with textual knowledge triplets, which is:

\begin{equation}
    V_T =\{s,o \mid \exists (s, r, o)\in T\}, E_T =\{(s_{t}, o_{t}, [r_{t}]) \mid \exists (s_{t}, r_{t}, o_{t})\in T\}
\label{eq:E_T}
\end{equation}
The formal representation is as follows: 

Note that we formulate the edge set $E$ as a edge $(s,o)$ with property $r$ to handle the relation between $s,o$.

The nodes in the visual subgraph $G_M$ are composed of the subjects or objects from the visual knowledge triplets $M$, and each visual knowledge triplet $m$ corresponds to a directed edge $e_m$ in the graph. The textual subgraph $G_T$ is constructed in the same manner. A concrete example is illustrated in Figure \ref{fig:method}.

However, answering a VQA question typically does not require utilizing all the multimodal information ($G_M$ and $G_T$) contained in the sample. Instead, only a subset of key information($G_K$) is necessary to arrive at the correct answer. Based on this observation, we extract the key knowledge triplets $K=\{(s_k, r_k, o_k))\} \subseteq M\cup T$ that are essential for answering the VQA question from the set of visual knowledge triplets $M$ and textual knowledge triplets $T$. These triples are then used to construct a new key multimodal knowledge subgraph $G_K = <V_K, E_K>$, which is:

\begin{equation}
    V_K =  \{s,o \mid \exists(s, r, o)\in K\}, E_K =\{(s_{k}, o_{k}, [r_{k}]) \mid \exists(s_{k}, r_{k}, o_{k})\in K\}
\end{equation}

The role of $G_K$ is similar to a graphical representation of the rationale required to answer a VQA question.

%G_K的作用就像是回答一个VQA问题对应rationale的图表示。

%目前我们就获得了对于一个VQA问题的图表示，VQA问题包含的潜在多模态知识信息被建模成视觉子图G_M和文本子图G_T，而回答一个VQA问题所需要使用到的关键信息被建模成为了关键子图G_K。而动态的改变VQA问题则体现为动态的改变VQA问题对应的关键子图G_K。
At this stage, we obtain a graph representation for each static VQA problem $S_0=\{I_0, Q_0, A_0\}$, $S_0$ denotes the original data of a given test VQA sample, where $I_0$, $Q_0$, and $A_0$ represent the corresponding input image, input question, and answer, respectively. The potential multimodal knowledge contained in the problem is modeled as a visual subgraph $G_M$ and a textual subgraph $G_T$, while the key information required to answer the question is captured by the key subgraph $G_K$. Consequently, dynamically altering a VQA problem can be naturally formulated as dynamically modifying its corresponding key subgraph $G_K$.	
 
\begin{equation}
    S_0 = \{I_0, Q_0, A_0\} \sim \{G_M, G_T, G_K\}
\end{equation}

%然而，回答一个VQA问题往往不需要使用到这个VQA sample中所包含的全部多模态信息，往往只有一部分关键信息是回答问题所需要的。由此我们从视觉知识三元组M和文本知识三元组T中抽取回答VQA问题所必须的关键知识三元组K，并根据这些三元组构建了新的关键多模态知识子图G_k

%视觉子图中的结点由视觉知识triplets中的s或者o组成。每一条视觉知识三元组构成了图中的每一条有向边。文本子图也是按照同样的方式构建而来，具体的示例可以参考图2
\paragraph{Graph Representation of Dynamic VQA} 
Intuitively, there are two ways to modify the key subgraph $G_K$ corresponding to a VQA sample.

\textbf{(1) Re-Selection}: by choosing a different set of key knowledge triples 
$K'=\{(s_{k'}, r_{k'}, o_{k'})\} \subseteq M\cup T \neq K$ from the existing visual subgraph $M$ and textual subgraph $T$, we can generate a new key subgraph $G_{K'} = <V_{K'}, E_{K'}>$. The formal expression is given as follows:
\begin{equation}
    V_{K'} = \{s,o \mid \exists(s, r, o)\in K'\}, E_{K'} =\{(s_{k^{'}}, o_{k^{'}}, [r_{k^{'}}])|\exists (s_{k^{'}}, r_{k^{'}}, o_{k^{'}})\in K'\}
\end{equation}

The new VQA sample $S_0'$ can be then formulated as:.
%此时，新的VQA问题S_0'可以如上表示。
\begin{equation}
    S_0' = \{I_0, Q_0', A_0'\} \sim \{G_M, G_T, G_{K'}\}
\end{equation}

\textbf{(2) External Knowledge Exploration}: by selecting appropriate knowledge triplets from external sources, we expand the original set of key triples $K$ into an extended set $K_n=\{(s_{k_n}, r_{k_n}, o_{k_n})\}$ with new textual triplets set $N$. Using $K_{n}$, we generate a new key subgraph $G_{K_n} = <V_{K_{n}},E_{K_{n}}>$. Unlike Re-Selection, $T_n$ is expanded together with the augmentation of the triplets. The formal expression is as follows:
\begin{equation}
    V_{K_n} = \{s,o \mid (s, r, o)\in K_n\}, E_{K_n} =\{(s_{k_{n}}, o_{k_{n}}, [r_{k_{n}}])\mid \exists(s_{k_{n}}, r_{k_{n}}, o_{k_{n}})\in K_n\}
\end{equation}

\begin{equation}
    T_n = N \cup T , K_n = N \cup  K.
\end{equation}

The new VQA sample $S_n$ can be formulated as shown above. 
\begin{equation}
    S_n = \{I_0, Q_n, A_n\} \sim \{G_M, G_{T_n}, G_{K_n}\}
\end{equation}

%与Re-Selection不同，M_n和T_n也会随着triplets的拓展二拓展,其中N_M是新增的视觉triplets，N_T则是新增的文本triplets，其形式如下所示:

As the corresponding graph structure is updated, the original VQA problem is transformed into a new one for evaluation. We view the difficulty of the generated VQA problem based on the number of edges $|E_K|$ in its key subgraph $G_K$.

%随着对应图结构的更新，原有的VQA问题也就会转化成为了新的VQA问题以供测试，我们使用最后生成的VQA问题中的关键子图G_K的边数来定义生成问题的难度。

%直观的，有两种方式可以用来改变一个VQA数据对应的关键子图G_K。(1)通过重新选择不同的关键知识三元组K从已有的视觉知识子图G_M和文本知识子图G_T中重新生成新的关键子图G_{K'}，其公式表现形式如下:

%(2)从外部知识来源挑选合适的知识三元组来拓展原有的知识三元组K到K1，从而结合原有的关键子图G_K来重新生成新的关键子图G_{K1}。其公式表现形式如下:

\subsection{Knowledge Enhanced Benchmark Evolution Framework}
%我们使用图的形式来表示VQA sample并建模动态评估过程，接下来会介绍我们的具体实现方式也就是our Dynamic Evaluation Framework。
We represent each VQA sample in the formulation of a graph and model the dynamic evaluation process accordingly. In the following, we present our concrete design, our Dynamic Evaluation Framework. Our overall pipeline can be divided into three components: Extract, Exploration, and express.

%我们整个流程可以分为Extract，Exploration和express三个部分。

\paragraph{Extract}
%我们首先基于一条已有的VQA数据$S_0$的输入图像，问题和答案进行信息提取，分别获取他的视觉知识三元组M和文本知识三元组T。然后，我们会基于提取的三元组结合原始VQA输入图像，问题与答案来选择其中的回答VQA问题所需要的关键三元组K。我们选择使用强大通用的多模态模型GPT4o来实现这一点。获取了对应的知识三元组M，T和K后，我们则会构建出原始数据VQA对应的图结构表示$G_M$, $G_T$与$G_K$。
We first perform information extraction based on the input image, question, and answer of the given VQA data, obtaining the corresponding visual knowledge triplets $M$ and textual knowledge triplets $T$. We then identify the key triplets $K$ that are required to answer the VQA question by combining the extracted triplets with the original VQA input. To achieve this, we employ the powerful and general-purpose multimodal model GPT-4o. Once $M$, $T$, and $K$ are obtained, we construct the graph representations of the VQA sample, namely  visual graph $G_M$,  textual graph $G_T$, and key subgraph $G_K$ according to Eq (\ref{eq:E_M}). We believe that for a reasonable VQA sample, its corresponding $G_K$ should contain at least one edge from $G_M$, i.e, at least one visual triplet. Otherwise, answering this VQA sample would not require any visual information, which we consider to be unreasonable. Therefore, we retain only results whose $G_K$ includes at least one edge originating from $G_M$. 

%我们认为对于一条合理的VQA sample而言，其所对应的G_K中至少应该包含一条来自于G_M的边，也就是至少包含一条视觉triplet。否则回答这条VQA数据不需要使用任务视觉信息，而这我们认为是不合理的VQA数据。因此，我们只保留了G_K中至少有一条边来源于G_M的这样的结果。

%The formal expression is as follows:

%S_0是对应的那一条测试VQA sample的原始数据，I_0, Q_0, A_0分别代表着对应VQA的输入图像，输入问题和对应的答案，G_M, G_T, G_K则是对于这一原始VQA问题的图表示。
% $S_0$ denotes the original data of a given test VQA sample, where $I_0$, $Q_0$, and $A_0$ represent the corresponding input image, input question, and answer, respectively. $G_M$, $G_T$, and $G_K$ denote the graph representations of this original VQA problem.

%exploration 
%为了保障拓展知识的可拓展性和准确性，我们对待拓展的Answer和Answer Triplets。

%为了保障拓展过程中知识的可靠性，我们需要引入fiilter过程。

\paragraph{Explore}
After obtaining $G_M$, $G_T$, and $G_K$ for an original question, we expand the original problem to generate new VQA questions. Specifically, we adopt two strategies for question expansion and generation: \textbf{Triplets Re-Selection} and \textbf{Triplets Exploration}.

To ensure the reliability of knowledge during the exploration process, we introduce a filtering step. We first perform an Answer filtering step, which consists of three components: representativeness filtering, part-of-speech filtering, and cycle-check filtering. Representativeness filtering is used to determine whether the corresponding triplet (s,r,o) is representative. Part-of-speech filtering examines the POS of the candidate Answer. Specifically, we assume that answers with a noun POS are more suitable for further exploration. Finally, cycle-check filtering ensures that for newly expanded triplets, the output cannot be identical to any subject in the original key triples, since this would introduce cycles in $G_K$, leading to unreasonble generated new question.

For a VQA sample $S_0$ from an existing dataset, we assume that representative filtering has already been considered during its construction, and thus all original questions are regarded as passing this step. We then apply part-of-speech filtering to the original answer $A_0$, which divides the samples into two groups: $Pos_T1$, where $A_0$	is a noun, and $Pos_F$, where it is not. For data belonging to $Pos_F$, we perform Re-Selection to construct new VQA questions whose answers are nouns, resulting in a new set $Pos_T2$. Finally, $Pos_T1$ and $Pos_T2$ together form $Pos_T$, the collection of VQA questions whose answers all satisfy the noun constraint. 

The concrete implementation of Re-Selection is as follows. We first identify the image root node of $G_M$, i.e., the node where v=IMAGE. We then search within $G_M$ and $G_T$ for paths that include this image root node. Among these, we define as the valid path set those paths whose terminal node (i.e., the endpoint of the last edge) is a noun. Finally, we select the longest path from this valid set to serve as the new key graph $G_{K'}$. A concrete example is illustrated in Figure \ref{fig:method}.

For the data in $Pos_T$, we perform knowledge exploration. Specifically, we first employ GPT to generate a set of candidate expandable knowledge triplets, where the subject s of each triplet corresponds to the current answer. We then apply our filtering strategies to these Answer-Related Triplets. Finally, from the filtered triplets that meet the requirements, we randomly select one for knowledge exploration and incorporate it into the current problem’s key subgraph $G_K$ to obtain new key subgraph $G_{K_1}$. Using the KBE-DME framework, we can iteratively repeat this process up to three times, thereby obtaining key subgraphs with different hop expansions, namely $G_{K_2}$ and $G_{K_3}$

%我们对Pos_T中的数据进行知识拓展，首先，我们使用GPT来生成可拓展的知识三元组候选集，其中这些三元组中的s是当前问题的Answer。然后我们会对这些待拓展的Answer Related Triplets应用我们的筛选策略，最后再从筛选后满足要求的triplets中随机选择一个进行知识拓展，将其加入到当前问题的G_K中。

%我们使用KBE-DME framework可以反复的迭代三次这一过程，从而得到不同跳数拓展的关键子图G_K2，G_K3。

%Re-Selection的具体实现方式如下：我们首先找到G_M中的图像根节点，也就是s为IMAGE的节点，然后我们会在G_M和G_T中搜索包含s的相关路径，我们选择路径的最后一条边的终结点为名词的路径为有效路径集，并选择其中长度最长的路径作为新的G_K，具体示例如图2所示。

%对于一条来自于已有数据集的VQA S_0而言，我们认为其在构建的时候就考虑到了代表性筛选，因此我们认为原始问题都能通过代表性筛选。我们对原始问题的答案A_0进行词性筛选，将原始问题根据A_0的词性分成两部分Pos_T1和Pos_F。其中Pos_T1是A_0为名词的，Pos_F则不是。我们对Pos_F的数据进行Re-Selection，构建出对应Answer为名词的新VQA问题Pos_T2。最后Pos_T1和Pos_T2共同组成Pos_T，即满足Answer均为名词的VQA问题。

%在获取了一个原始问题的G_M, G_T, G_K后，我们会对原始问题进行拓展并生成新的VQA问题。我们分别会采用Triplets Re-Selection和Triplets Exploration两种方式来进行新问题的拓展生成。

%整体流程如下，我们首先会对问题的Answer进行筛选，筛选过程包括代表性筛选，词性筛选和环判断筛选。其中代表性筛选是用来判断对应的triplet的(s, r, o)是否具有代表性，词性筛选则是判断对应Answer的词性，我们认为对应Answer词性为名词时方便直接进行拓展，环判断筛选则是对于新拓展的triplets，其对应的output不能与原有key_triplets中的某一个s相同，否则会在G_K中产生回路，影响难度控制。

\paragraph{Express}
We employ GPT to transform the generated key subgraph into a new VQA question–answer pair. Taking the first expansion as an example, we provide GPT with the image, question, and answer of the current VQA sample $S_0$, along with its corresponding key subgraph $G_K$. We then specify the knowledge triplet to be expanded and designate the new VQA answer as the output of this triplet. Finally, GPT is instructed to generate a new input question based on this information, thereby completing the transformation from the graph representation to a new VQA sample.
%我们采用GPT将生成的关键子图转换为新的VQA QA对。以第一次拓展举例，我们会提供当前VQA S_0的图像，问题和答案，并给出当前问题的关键子图G_K，然后我们会指定拓展的知识triplet，并指定新VQA问题的Answer为对应制定triplet的output。最后要求GPT根据这些信息来生成新VQA问题的输入问题，从而完成从图到VQA sample的转换。

%考虑到alias情况，我们使用GPT来判断测试模型在VQA问题的回复与提供的Answer是否对应。

\section{Experiment}
\label{sec:Experiment}
\subsection{Experiment Setup}
\paragraph{Datasets}
We choose OK-VQA\citep{marino2019ok} and A-OKVQA\citep{schwenk2022aokvqabenchmarkvisualquestion} as the primary static datasets for our experiments. Specifically, we select the validation splits of these datasets. The validation sets of OK-VQA and A-OKVQA contain approximately 5k and 1.1k samples, respectively. From these, we select 2.6k samples from OK-VQA and 0.6k samples from A-OKVQA as the starting points for the static test sets in our dynamic evaluation. 

%我们选择OK-VQA和A-OKVQA这两个数据集来作为我们实验的主要静态数据集来源。我们选择了这两个数据集的validation作为静态测试的VQA Benchmark。OK-VQA和A-OKVQA的验证集分别拥有5k和1.1k的测试数据。我们分别选择了其中的2.6k和0.6k数据来作为动态评估中的静态测试集起点。

%我们分别选择了

\paragraph{Evaluated MLLMs}
%我们的测试模型包括闭源API测试模型：GPT-4o, Gemini-2.5-pro, Claude和开源的待测模型LLaVA-OV和Qwen-2.5-VL。为了较为公平的回答VQA问题，我们约束测试模型的回复不能过长。
Our evaluation covers both closed-source models, including GPT-4o\citep{openai2024gpt4ocard}, Gemini-2.5-pro\citep{comanici2025gemini25pushingfrontier}, and Claude\citep{anthropic_claude3_card}, as well as open-source models und, namely LLaVA-OV-7B\citep{li2024llavaonevisioneasyvisualtask} and Qwen-2.5-VL-7B\citep{bai2025qwen25vltechnicalreport}. To ensure fair comparison in answering VQA questions, we restrict the length of the models’ responses. Considering possible alias cases, we employ GPT-4o to determine whether the response of the tested model to a VQA question corresponds to the provided answer.

\subsection{Main Results}
%我们对原始问题按照图2的流程进行了最多3次的拓展，我们在不同长度拓展后的数据上测试了5个多模态大模型的效果，结果如表所示。
We expand each original question up to three hops following the procedure illustrated in Figure 2. We then evaluate five multimodal large language models on the datasets obtained after expansion of different hops, with the results presented in the Table \ref{tab:main_performance}.

We observe that as the number of expansion hops increases, the performance of all five tested models declines across both datasets. This indirectly demonstrates that our dynamic evaluation framework provides reliable control over task difficulty. 

In addition, we find that some models exhibit relatively smooth performance degradation across different expansion hops, such as GPT-4o, Claude, and Qwen-2.5-VL in the dynamic evaluation based on the OK-VQA dataset. However, for Gemini-2.5-pro and LLaVA-OV, the performance drop is more pronounced during the first expansion, while the decline becomes more gradual in subsequent hops. In the dynamic evaluation based on the A-OKVQA dataset, GPT-4o and Claude again maintain relatively smooth degradation, whereas Qwen-2.5-VL shows a comparatively larger drop at the first expansion than at later ones.

The marginal effect makes it reasonable that the performance gap between models diminishes as the test questions become more difficult. However, if a model exhibits a substantial performance drop at the very first expansion, this may indicate a potential risk of data contamination on the corresponding dataset.

%我们可以看到，随着拓展跳数的增加，5个测试模型在两个数据集上的表现都在下降，这也在侧面反映了我们的动态评估框架的难度控制是有保障的。

%另外，我们发现一些模型在不同的跳数拓展得到的VQA问题上的performance下降比较平滑，诸如在基于OK-VQA数据集的动态评估中GPT-4o，Claude和Qwen-2.5-VL。但是对于Gemini-2.5-pro和LLaVA-OV来说，这些模型在首次拓展时的性能下降更为明显，在后续的拓展中，性能下降趋势反而放缓。而在基于A-OKVQA数据集的动态评估中，GPT-4o和Claude依旧保持较为平滑的性能下降，然而Qwen-2.5-VL出现了与其他拓展相比第一跳拓展性能下降较大的情况。

%边际效应导致模型在测试越难的问题之间的性能差别降低是合理的，然而如果模型在首次拓展时就发生较大的性能损失，那么此模型在对应数据集上可能存在数据污染的风险。

\begin{table}[tp]
%\begin{center}
\centering  
\setlength{\tabcolsep}{2.5mm}{
\renewcommand\arraystretch{1.3}
\caption{Main Results of five different MLLMs on original static benchmark(raw) and our generated dynamic benchmark with exploration of different(1-3) hops through our KBE-DME framework.}
\label{tab:main_performance}
\begin{tabular}{l|llll|llll}
\toprule[1pt]
\multirow{2}{*}{\bf Model}  &\multicolumn{4}{c|}{\bf OK-VQA} & \multicolumn{4}{c}{\bf A-OKVQA}\\ \cline{2-9}
                            & Raw   & 1-hop  & 2-hop  & 3-hop  & Raw   & 1-hop  & 2-hop  & 3-hop
\\ \midrule[1pt] 

GPT-4o                      & 50.33 & 47.62 & 42.07 & 39.19 & 60.13 & 50.82 & 41.34 & 36.27\\
Gemini-2.5-pro              & 49.94 & 41.55 & 36.87 & 32.92 & 58.99 & 46.57 & 34.48 & 32.03\\
Claude                      & 52.26 & 46.03 & 41.82 & 37.60 & 57.68 & 48.85 & 38.73 & 32.84\\
LLaVA-OV                    & 53.46 & 36.53 & 31.45 & 29.52 & 60.78 & 38.56 & 30.56 & 25.33\\
Qwen-2.5-VL                 & 47.35 & 42.24 & 37.68 & 33.04 & 57.35 & 43.95 & 37.42 & 30.88\\
\bottomrule[1pt]
\end{tabular}}
%\end{center}
\end{table}

\begin{table}[tp]
%\begin{center}
\centering  
\setlength{\tabcolsep}{2.5mm}{
\renewcommand\arraystretch{1.3}
\caption{Several statistical metrics of original VQA data and the VQA questions generated with exploration of different hops. The statistical metrics including the average number of words in the questions, the average number of words in the answers, the average number of edges $|E_K|$ in the key subgraphs $G_K$, and the number of distinct relations among the triplets in the entire corresponding dataset.}
\label{tab:statstics}
\begin{tabular}{l|llll|llll}
\toprule[1pt]
\multirow{2}{*}{\bf Attribute}  &\multicolumn{4}{c|}{\bf OK-VQA} & \multicolumn{4}{c}{\bf A-OKVQA}\\ \cline{2-9}
                            & Raw   & 1-hop  & 2-hop  & 3-hop  & Raw   & 1-hop  & 2-hop  & 3-hop
\\ \midrule[1pt] 

Question Words              & 8.18  & 15.2   & 17.5   & 18.8   & 8.89  & 15.7   & 17.9   & 19.3 \\
Answer Words                & 1.19  & 1.50   & 1.53   & 1.53   & 1.17  & 1.45   & 1.54   & 1.52 \\
$|E_K|$                     & 2.98  & 4.04   & 5.04   & 6.04   & 3.16  & 4.23   & 5.23   & 6.23 \\
All Relations               & 2979  & 4172   & 5136   & 6011   & 1263  & 1602   & 1900   & 2175 \\
\bottomrule[1pt]
\end{tabular}}
%\end{center}
\end{table}

%原始VQA数据和不同跳数拓展生成的VQA问题的统计指标，我们统计了这些VQA sample中的平均问题单词数，平均答案单词数，关键子图中的平均边数和整个数据集涉及到的triplets的relation种类数。

%Several statistical metrics of original VQA data and the VQA questions generated with different expansion hops. The statistical metrics including the average number of words in the questions, the average number of words in the answers, the average number of edges in the key subgraphs, and the number of distinct relation types among the triplets in the entire dataset.

\section{Quality Analysis}
\label{sec:Quality}
\subsection{Statistics}
%为了更好的分析我们新生成数据和原始VQA数据的区别，我们对新老VQA数据进行了一些统计学指标上的分析。结果如表所示:
To better analyze the differences between our newly generated data and the original VQA data, we conduct a statistical analysis on both sets of VQA samples. The results are presented in Table \ref{tab:statstics}. 
We observe that on both benchmarks, as the number of expansion hops increases, the newly generated VQA samples tend to have longer questions. This is often due to the fact that answering these questions requires longer reasoning chains, which in turn makes the questions more complex. We also find that the number of edges in the key subgraph increases steadily with more hops, and consequently, the number of relation types involved in the visual and textual knowledge triples also grows. These results demonstrate that the newly generated VQA questions achieve higher distributional diversity, greater question complexity, and longer rationales compared to the original or lower-hop expansions, thereby producing more challenging VQA problems. This suggests that through the KBE-DME framework, we can evolve a static VQA benchmark into a dynamically changing dataset with controllable difficulty for the dynamic evaluation of multimodal large language models.

%可以发现，随着跳数的增加，新生成的VQA sample往往具有更长的问题，这往往是由于问答VQA所需要的推理链路变长，因此VQA的问题也会变得更加复杂。可以看到关键子图中的边数在随着跳数增加而稳定增加，由此生成的VQA数据所对应的视觉知识三元组和文本知识三元组中涉及到的关系种类也随之增加。这些都可以证明，新生成的VQA问题在分布多样性上，问题复杂程度上，以及问答问题所需要的rationale长度都得到了提升，从而生成比原始问题或者上一跳拓展的问题更难的VQA问题。通过KBE-DME framework,我们可以使得原始静态VQA benchmark进化为难度可控的动态变化的VQA数据集，来对多模态大模型进行动态化的评估。

\subsection{Human Study}
In addition to ensuring the diversity of dynamically generated data with controllable difficulty, it is also essential to guarantee their quality. We sampled 150 dynamically generated evaluation VQA instances from the OK-VQA dataset for human evaluation. We assessed the dynamic generation process of VQA questions from three perspectives: (1) whether the newly generated VQA sample is itself reasonable as a VQA problem (VQA\_Reasonable); (2) whether each triplet in the corresponding set of key knowledge triples $K$ is correct (Triplet\_Correct); and (3) whether the generated VQA question is consistent with its corresponding key knowledge triplets (VQA\_Triplets\_Alignment). The evaluation results are presented in the Table \ref{tab:human}. Prompt can be seen in Appendix \ref{sec:Prompts}. 
%\red{More Details can be seen in Appendix.}
The human evaluation results demonstrate that our dynamic evaluation framework, KBE-DME, can generate high-quality VQA data with correct and well-aligned key triplets. This further validates the accuracy of our framework and the reliability of the generated VQA data.

%除了保障难度可控动态生成数据的多样性，还需要保障动态生成数据的准确性。我们从OK-VQA数据集中抽取了150条生成的动态评估数据进行人类评价。我们从三个角度来评价VQA问题的动态生成过程。首先是评价新生成的VQA数据本身作为一个VQA问题是否合理，即VQA Reasonable。其次是判断当前VQA问题对应的关键知识三元组K中的每个triplet是否正确，即Triplet_Correct。最后是判断生成的VQA问题与对应的关键知识三元组是否一致，即VQA Triplets Alignment。测试结果如表所示:

%我们在抽取的150条生成的VQA问题中，检查了新生成的VQA数据的(1)VQA\_Reasonable: 是否是一个合理的VQA sample. (2)Triplets_Correct: 其中的Key Triplets中的每个triplet的内容是否正确。(3)VQA_Triplets_Alignment: VQA问题与Key Triplets是否对应。我们展示了这些问题的平均人类评价情况，括号中的是不同的标注者的一致性。

%人工评价的结果证明我们的动态评估框架KBE-DME可以生成高质量的VQA数据，并且相对应的key triplets是正确和对齐的。同时证验证了我们framework的准确性和生成的VQA数据的可靠性。

\begin{table}[tp]
%\begin{center}
\centering  
\setlength{\tabcolsep}{3mm}{
\renewcommand\arraystretch{1.3}
\caption{Human\_Study of KBE-DME on OK-VQA Benchmark. For the 150 generated VQA samples, we evaluated: (1) VQA\_Reasonable: whether the sample is a reasonable VQA problem; (2) Triplets\_Correct: whether each key triplet is correct, and (3) VQA\_Triplets\_Alignment: whether the question aligns with the key triplets. We report the average human evaluation scores for these generated questions, with the values in parentheses indicating inter-annotator agreement.}
\label{tab:human}
\begin{tabular}{lllll}
\toprule[1pt]
                & VQA\_Reasonable & Triplets\_Correct & VQA\_Triplets\_Alignment\\ \midrule[1pt] 
3-hop Aver             & 95.0(90.1\%)  & 96.8(93.6\%)    & 97.9(95.7\%)\\
\bottomrule[1pt]
\end{tabular}}
%\end{center}
\end{table}

% VQA\_type        & VQA\_Reasonable & Triplets\_Judge & VQA\_Triplets\_Alignment\\ \midrule[1pt] 
% Raw              & 85.1(78.7\%)    & 94.7(89.4\%)    & 97.9(95.7\%)   \\
% 1-hop            & 94.7(89.4\%)    & 94.7(89.4\%)    & 97.9(95.7\%)   \\
% 2-hop            & 94.7(89.4\%)    & 97.9(95.7\%)    & 97.9(95.7\%)   \\
% 3-hop            & 95.7(91.5\%)    & 97.9(95.7\%)    & 97.9(95.7\%)   \\

\begin{figure}[t]
\begin{center}
%\framebox[4.0in]{$\;$}
    \includegraphics[width=0.9\linewidth]{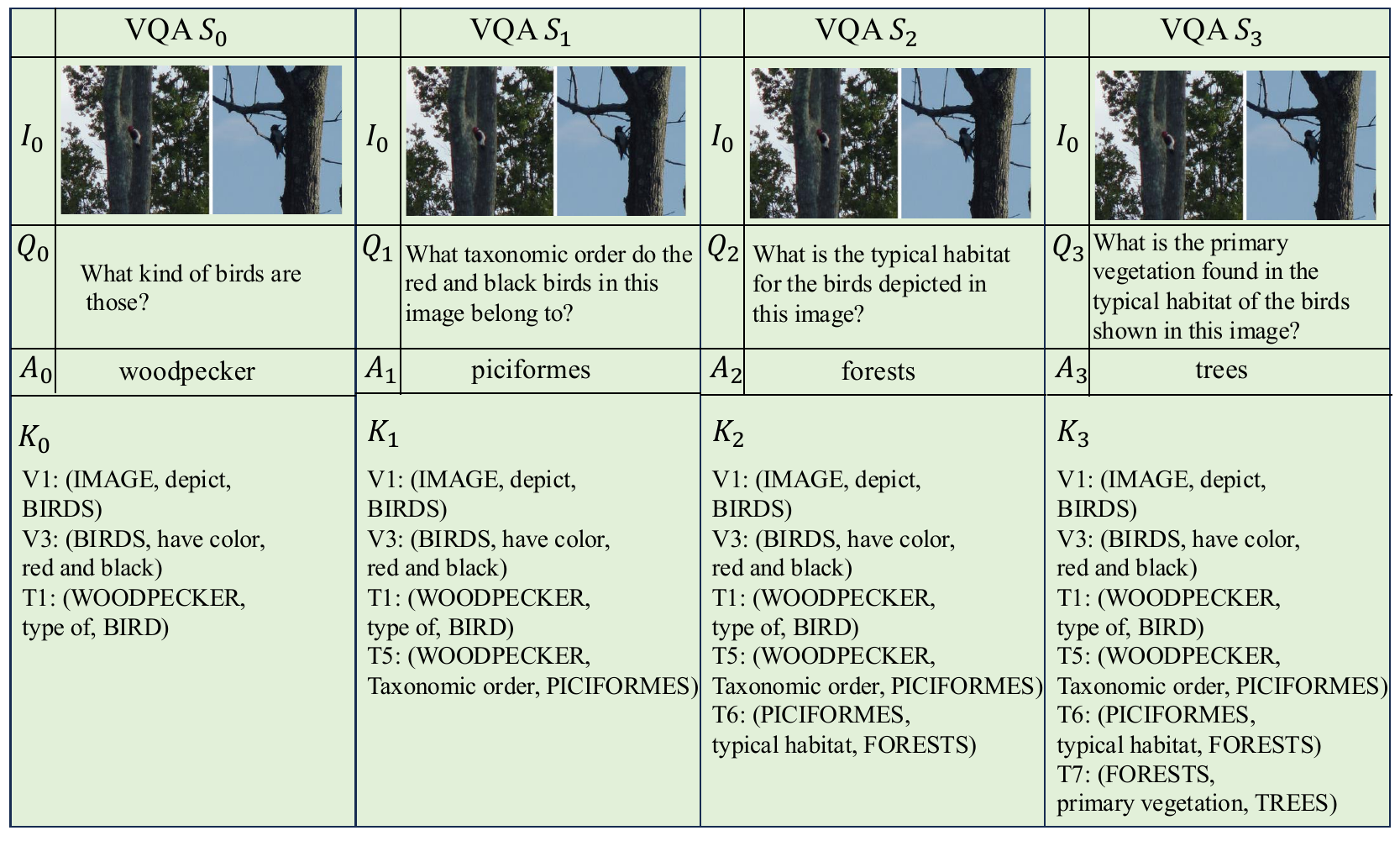}
    \caption{An example from our data construction process on OK-VQA. $S_0$ denotes the original VQA sample, where $I_0$, $Q_0$, and $A
_0$	represent the corresponding image, question, and answer, and $K_0$ is the associated set of key knowledge triples. For the generated data $S_i$ at different hop levels i, we keep the input image unchanged while reconstructing the corresponding questions and answers by altering the composition of the key knowledge triples.}
    \label{fig:case_study}
\end{center}
\end{figure}

%这里是我们数据构造中的一个例子，S_0是原始的VQA数据，I_0, Q_0, A_0分别代表对应的图像，问题和答案，K_0是其对应的相关关键知识三元组信息。在不同跳数i的生成数据S_i中，我们保持输入图像不变，通过改变关键知识三元组的组成信息来重新构建相应的问题和答案。三元组的编号是其在对应视觉/文本子图中的编号。

\subsection{Case Study}
We present an example of the data construction process from OK-VQA, as illustrated in the Figure \ref{fig:case_study}. Our KBE-DME first analyzes the original VQA problem and identifies the key knowledge triples. After the filtering process and the possible Re-Selection step (as illustrated in the Figure\ref{fig:method}), KBE-DME searches for textual knowledge triples whose subjects correspond to the original answer (such as T5: (WOODPECKER, Taxonomic order, PICIFORMES)). Following an additional filtering step, a selected triple is incorporated into the original key knowledge set, and based on this updated set of key triples, KBE-DME generates a new VQA question as "What taxonomic order do the red and black birds in this image belong to?". By repeating this process, we can iteratively expand and generate new VQA question–answer pairs corresponding to different sets of key knowledge triples, indicating the effectiveness of our proposed framework.

%我们展示了一个来自OK-VQA的数据构造流程，如图所示:

%可以发现我们的KBE-DME首先会分析原始VQA问题，并找到与之相关的关键知识三元组信息。经过如图所示的筛选以及可能的Re-Selection过程后，KBE-DME会基于当前的VQA问题寻找原始Answer为s的文本知识三元组，对找到的这些三元组经过筛选后，KBE-DME会将选择的三元组加入到原来的关键知识三元组中，然后根据新的关键知识三元组生成新的VQA问题。

%通过多次执行这一过程，我们可以迭代式的拓展生成出由不同关键知识三元组对应的新VQA问题答案对。

% \begin{table}[htp]
% %\begin{center}
% \centering  
% \setlength{\tabcolsep}{2.5mm}{
% \renewcommand\arraystretch{1.2}
% \caption{Main Performance using Gemini}
% \label{main_performance}
% \begin{tabular}{l|llll|llll}
% %\renewcommand{\arraystretch}{1.1}
% %\renewcommand{\arraystretch}{1.3}
% \toprule
% \multirow{2}{*}{\bf Model}  &\multicolumn{4}{c|}{\bf OKVQA} & \multicolumn{4}{c}{\bf AOKVQA}\\ \cline{2-9}
%                             & Raw   & 1hop  & 2hop  & 3hop  & Raw   & 1hop  & 2hop  & 3hop
% \\ \midrule 

% GPT-4o                      \\
% Gemini-2.5-pro              \\
% \bottomrule
% \end{tabular}}
% %\end{center}
% \end{table}

\section{Conclusion}
\label{sec:Conclusion}
%CONCLUSION AND DISCUSSION
%在这篇论文里，我们首先使用图表示方法来表示VQA测试数据和动态评估过程，然后提出了KBE-DME动态评估framework来对模型进行动态评估。我们基于OK-VQA和A-OKVQA两个静态VQA benchmark动态构建了不同难度等级的VQA测试数据来进行动态评估。我们对动态构建的数据多样性分布和质量进行了分析，并使用这些数据动态评估了5个不同的开源或闭源多模态大模型在不同难度VQA测试数据下的表现。实验证明KBE-DME框架可以动态可变的生成难度可控的高质量测试数据来进行动态评估，而评估结果展示所有的测试模型在更难的数据上的表现下降，以及部分模型潜在的数据污染风险。KBE-DME能够广泛的使用到不同的多模态Benchmark上，并解决传统静态评估所存在的数据饱和和数据污染风险。
In this paper, we first introduce a graph-based representation to model VQA data and the dynamic evaluation process. We then propose KBE-DME, a dynamic multimodal evaluation framework. Building upon two static VQA benchmarks, OK-VQA and A-OKVQA, we dynamically construct VQA test samples with varied difficulty levels and conduct comprehensive analyses of the diversity and quality of our dynamically constructed data. We further evaluate five different open- and closed-source multimodal large language models under these dynamically generated test data. Experimental results show that KBE-DME can dynamically generate high-quality test data with controllable difficulty, while the evaluation results reveal consistent performance degradation of all tested models on harder data. Overall, KBE-DME provides a generalizable framework that can be applied to diverse multimodal benchmarks, effectively alleviates the risk of data saturation and contamination inherent in traditional static evaluation.

\section*{Ethics Statement}
We adhere to the ICLR Code of Ethics. Our released dataset sources data from open-source datasets as indicated, following their license and copyright restrictions. Our released dataset containing synthetic data is for research only and does not aim at conveying any information about real-life.

\section*{Reproducibility statement}
We submit our dataset through supplementary material. We have disclosed the models and prompts used for data generation in Section \ref{sec:KBE-DME} and \ref{sec:Experiment}. We have clearly cited and listed the checkpoints of the MLLMs used for evaluation in \ref{sec:Experiment}.

\bibliography{iclr2026_conference}
\bibliographystyle{iclr2026_conference}

\appendix
\section{Prompt}
\label{sec:Prompts}
The prompts are presented as follows:

\begin{figure}[htbp]
        \centering
        \begin{tcolorbox}[title=Graph Extraction Prompt]
        You are a helpful assistant. You need to analyze the visual information subgraph and textual information subgraph implied in the input information of a VQA instance, which includes an image, a question, and an answer. Please output and number the visual and textual knowledge subgraphs in the form of knowledge triples. Then please generate the corresponding answer rationale in the form of knowledge triples, including only the necessary knowledge triples.

Here is an example of output visual and textual knowledge subgraphs.

Visual Information Subgraph:

V1.(Image, contains, motorcycle)

V2.(motorcycle, has color, black)

V3.(motorcycle, has component, engine)

V4.(motorcycle, has feature, two wheels)

V5.(motorcycle, is on, road or track)

Textual Information Subgraph:

T1.(motorcycle, can be used for, race)

T2.(sport, has type, race)

T3.(race, requires, high speed vehicle)

T4.(high speed vehicle, includes, motorcycle)

Multimodal Answer Rationale:

V1.(Image, contains, motorcycle)

T1.(motorcycle, can be used for, race)

T2.(sport, has type, race)

Now, please generate the visual and textual information subgraphs for a VQA example.

Here is the VQA instance:

Image: The given image

Question: \{Question content\}

Answer: \{Model response\}

Please generate: the visual knowledge subgraph implied in the image

and the textual knowledge subgraph implied in the question and answer.

Visual Information Subgraph:

Textual Information Subgraph:

Multimodal Answer Rationale:
        \end{tcolorbox}
\caption{Graph extraction prompt.}
\label{fig:image-caption-prompt}
\end{figure}

\begin{figure}[htbp]
        \centering
        \begin{tcolorbox}[title=Key Triplets Extraction Prompt]
You are a helpful assistant. You will be given a (VQA) Visual Question Answering instance that includes an input image, a question, and its corresponding answer, as well as a set of corresponding visual and textual information in the form of triplets. The triplets in the visual information contain only visual information implied in the image, excluding any textual background knowledge. The triplets in the textual information include only relevant textual background knowledge.
Please select the necessary key information triplets that are required to answer this VQA question. Below are some examples:

======Example1======:

VQA Question: The man wearing a hat what is the name of that hat?

VQA Answer: cowboy hat

Visual Information triplets:

V1: (IMAGE, depict, MAN)

V2: (MAN, wear, HAT)

V3: (HAT, have type, COWBOY HAT)

V4: (MAN, ride, HORSE)

V5: (HORSE, is on, PATH)

V6: (PATH, is in, MOUNTAINOUS AREA)

V7: (IMAGE, contain, BACKPACK)

V8: (BACKPACK, have color, red)

Textual Information triplets:

T1: (COWBOY HAT, is a type of, HAT)

T2: (COWBOY HAT, typically worn by, COWBOYS)

T3: (COWBOY, commonly associated with, HORSE RIDING)

T4: (COWBOY HAT, used for, SUN PROTECTION)

Key information triplets to answer the question:

V1: (IMAGE, depict, MAN)

V2: (MAN, wear, HAT)

V3: (HAT, have type, COWBOY HAT)

======Example2======:

VQA Question: How many teeth does this animal use to have?

VQA Answer: 26

Visual Information triplets:

V1: (IMAGE, depict, CAT)

V2: (CAT, have color, beige)

V3: (CAT, is on, WINDOWSILL)

V4: (WINDOWSILL, is part of, WINDOW)

V5: (CAT, is in state, RELAXED)

Textual Information triplets:

T1: (ANIMAL, typically have, TEETH)

T2: (CAT, category of, ANIMAL)

T3: (CAT, usually have, 26 TEETH)

Key information triplets to answer the question:

V1: (IMAGE, depict, CAT)

T3: (CAT, usually have, 26 TEETH)

Now, please select the Key information triplets given the image, question and answer of a VQA example with its corresponding Visual and Textual Information triplets.

======VQA Input======

Image: The given image

VQA Question: \{VQA\_Q\}

VQA Answer: \{VQA\_A\}

Visual Information triplets:

\{Visual\_Information\_triplets\_str\}

Textual Information triplets:

\{Textual\_Information\_triplets\_str\}

Key information triplets to answer the question:
        \end{tcolorbox}
\caption{Key triplets extraction prompt.}
\label{fig:image-caption-prompt}
\end{figure}

\begin{figure}[htbp]
        \centering
        \begin{tcolorbox}[title=Knowledge Generation Prompt]
You are a helpful assistant. You will receive a VQA example. Please generate some knowledge triplets related to the answer. You should understand the meaning of the answer by combining the image and the question in the VQA, and then generate answer-related knowledge triplets. The newly generated knowledge triplets should not conflict with the information in the original VQA example. A triplet can be represented as (s, r, o), where s is the subject (an object), r is the corresponding relation, and o is either an attribute of the object or another object. Use uppercase for objects and lowercase for attributes. Here, s is the relation subject, r is the related relation, and o is the result of the relation corresponding to s. Generated knowledge triplets (s, r, o) should follow the following requirements:

1.The subject (s) must always be the answer itself.

2.The relation (r) must be specific and unique. Do not use vague terms like is a or has; instead, refine them into clear categories or attributes, such as taxonomic\_class, primary\_covering, foot\_type.

3.Please ensure that within a triplet (s, r, o), the object (o) is unique given the specified subject (s) and relation (r). In a triplet (s, r, o), the o must be an object, not an attribute.

4.The output format must strictly be one triplet per line: (s, r, o).

Below are some examples:

=========================Example1===========================
VQA\_Question: What country does this appear to be?

VQA\_Answer: scotland

Answer Related Knowledge Triples:

(SCOTLAND, geographic\_location, united\_kingdom)

(SCOTLAND, primary\_landscape, highlands)

(SCOTLAND, common\_tree\_type, deciduous)

=========================Example2===========================
VQA\_Question: What animal is this boat mimicing?

VQA\_Answer: duck

Answer Related Knowledge Triples:

(DUCK, taxonomic\_class, AVES)

(DUCK, taxonomic\_order, ANSERIFORMES)

(DUCK, taxonomic\_family, ANATIDAE)

(DUCK, common\_category, WATERFOWL)

(DUCK, typical\_habitat, WATER)

(DUCK, primary\_covering, FEATHERS)

(DUCK, mouth\_structure, BEAK)

(DUCK, foot\_type, WEBBED\_FEET)

(DUCK, typical\_sound, QUACK)

(DUCK, diet\_type, OMNIVORE)

Below is the VQA sample for generating extended knowledge:

Image: The given image

VQA\_Question: \{VQA\_Q\}

VQA\_Answer: \{VQA\_A\}

After generating the relevant (s, r, o) triplets, check each triplet individually and generate all possible o values for the given s and r. If a tuple (s, r, o) contains multiple outputs for the same s and r, delete that tuple. If o is an attribute rather than an object, also delete that tuple.

Answer Related Knowledge Triples:
        \end{tcolorbox}
\caption{Knowledge generation prompt.}
\label{fig:image-caption-prompt}
\end{figure}

\begin{figure}[htbp]
        \centering
        \begin{tcolorbox}[title=Knowledge Generation Prompt]
You are a helpful assistant. You will receive a VQA example. Please generate some knowledge triplets related to the answer. You should understand the meaning of the answer by combining the image and the question in the VQA, and then generate answer-related knowledge triplets. The newly generated knowledge triplets should not conflict with the information in the original VQA example. A triplet can be represented as (s, r, o), where s is the subject (an object), r is the corresponding relation, and o is either an attribute of the object or another object. Use uppercase for objects and lowercase for attributes. Here, s is the relation subject, r is the related relation, and o is the result of the relation corresponding to s. Generated knowledge triplets (s, r, o) should follow the following requirements:

1.The subject (s) must always be the answer itself.

2.The relation (r) must be specific and unique. Do not use vague terms like is a or has; instead, refine them into clear categories or attributes, such as taxonomic\_class, primary\_covering, foot\_type.

3.Please ensure that within a triplet (s, r, o), the object (o) is unique given the specified subject (s) and relation (r). In a triplet (s, r, o), the o must be an object, not an attribute.

4.The output format must strictly be one triplet per line: (s, r, o).

Below are some examples:

=========================Example1===========================
VQA\_Question: What country does this appear to be?

VQA\_Answer: scotland

Answer Related Knowledge Triples:

(SCOTLAND, geographic\_location, united\_kingdom)

(SCOTLAND, primary\_landscape, highlands)

(SCOTLAND, common\_tree\_type, deciduous)

=========================Example2===========================
VQA\_Question: What animal is this boat mimicing?

VQA\_Answer: duck

Answer Related Knowledge Triples:

(DUCK, taxonomic\_class, AVES)

(DUCK, taxonomic\_order, ANSERIFORMES)

(DUCK, taxonomic\_family, ANATIDAE)

(DUCK, common\_category, WATERFOWL)

(DUCK, typical\_habitat, WATER)

(DUCK, primary\_covering, FEATHERS)

(DUCK, mouth\_structure, BEAK)

(DUCK, foot\_type, WEBBED\_FEET)

(DUCK, typical\_sound, QUACK)

(DUCK, diet\_type, OMNIVORE)

Below is the VQA sample for generating extended knowledge:

Image: The given image

VQA\_Question: \{VQA\_Q\}

VQA\_Answer: \{VQA\_A\}

After generating the relevant (s, r, o) triplets, check each triplet individually and generate all possible o values for the given s and r. If a tuple (s, r, o) contains multiple outputs for the same s and r, delete that tuple. If o is an attribute rather than an object, also delete that tuple.

Answer Related Knowledge Triples:
        \end{tcolorbox}
\caption{Knowledge generation prompt.}
\label{fig:image-caption-prompt}
\end{figure}

\begin{figure}[htbp]
        \centering
        \begin{tcolorbox}[title=Representative Filter Prompt]
Here are some triplets related to a VQA example. Each triplet is composed of (s, r, o). I will provide you with the corresponding VQA example, and based on the VQA context, you need to determine whether the given o is representative for the specified s and r.

If it is representative, please output Yes.

If the relation is too broad or ambiguous to determine a unique representative, simply output No. 

Please output only the triplet numbers and their corresponding results: Yes or No.

You must evaluate every triplet and output the corresponding result in order.

Here are some examples:

==========Example1==========

VQA\_Question: What type of platform should this vehicle be on?

VQA\_Answer: track

Related Triplets:

1.(TRACK, primary\_use, TRANSPORTATION) 

2.(TRACK, common\_association, TRAINS) 

3.(TRACK, typical\_material, STEEL)

Representative Judgment:

1.No

2.Yes

3.Yes

==========Example2==========

VQA\_Question: Name the material used to make this car seat shown in this picture? 

VQA\_Answer: cloth

Related Triplets: 

1.(CLOTH, typical\_use, UPHOLSTERY) 

2.(CLOTH, material\_origin, TEXTILE) 

3.(CLOTH, common\_source, PLANT\_FIBERS)

Representative Judgment:

1.Yes

2.No

3.Yes

The following are examples to be judged:

Image: The given image.

VQA\_Question: \{VQA\_Q\}

VQA\_Answer: \{VQA\_A\}

Related Triplets: 
\{Related\_Triplets\}

Representative Judgment:
        \end{tcolorbox}
\caption{Representative filter prompt.}
\label{fig:image-caption-prompt}
\end{figure}

\begin{figure}[htbp]
        \centering
        \begin{tcolorbox}[title=Question Generation Prompt]
Please generate a new VQA question based on an original VQA question and the related information triplets. A triplet can be represented as (s, r, o), where s is the subject (an object), r is the corresponding relation, and o is either an attribute of the object or another object. Use uppercase for objects and lowercase for attributes. We will provide you with the triplets necessary for forming the new question. These triplets consist of those used to answer the original VQA question as well as additional newly introduced triplets.
We will also specify the answer for the new question along with the corresponding answer information triplet. You should combine the given original question and all the knowledge triplets to generate the new VQA question.
The new VQA question must ensure that its answer is the specified one and that it is related to the provided answer triplet. The new question must not contain the original question’s answer, and it should require the use of all provided knowledge triplets in order to be answered. Apart from the information in the newly added triplets, the new question must not include more information than the original question. Below are some examples:

======Example======:

Original VQA Question: What country does this appear to be?

Original VQA Answer: scotland.

Related Information Triplets for Original VQA sample:

visual\_triplets\_list:

V2: (IMAGE, depict, SHEEP)

V3: (IMAGE, depict, LAND ROVER)

textual\_triplets\_list:

T1: (SHEEP, commonly found in, SCOTLAND)

T2: (LAND ROVER, associated with, BRITISH COUNTRYSIDE)

T3: (BRITISH COUNTRYSIDE, includes, SCOTLAND)

Related Information triplet for New Answer in Generated VQA sample:

(SCOTLAND, traditional\_clothing, KILT)

New Answer in Generated VQA sample:
KILT

Generated new VQA Question:
What is the traditional clothing of the country shown in this image?

Please generate a new VQA question based on the information provided below, following the given example and requirements. The provided information is as follows:

Original VQA Image: The given image.

Original VQA Question: \{Ori\_VQA\_Q\}

Original VQA Answer: \{Ori\_VQA\_A\}.

Related Information Triplets for Original VQA sample:
\{Key\_Triplets\_str\}

Related Information triplet for New Answer in Generated VQA sample:
\{New\_VQA\_related\_Triplets\}

New Answer in Generated VQA sample:
\{New\_VQA\_Answer\}

Generated new VQA Question:
        \end{tcolorbox}
\caption{Question generation prompt.}
\label{fig:image-caption-prompt}
\end{figure}

\begin{figure}[htbp]
        \centering
        \begin{tcolorbox}[title=Judging Prompt]
Please analyze whether a given response to a VQA question matches its corresponding answer. If they match, output "Yes"; otherwise, output "No". Only output the judgment result "Yes" or "No". We will provide relevant image information to assist in the judgment.

Image: The given image.

Response: \{Response\}

Answer: \{Answer\}
        \end{tcolorbox}
\caption{Judging prompt.}
\label{fig:image-caption-prompt}
\end{figure}

\section{Use of LLMs}
LLMs are employed to facilitate our writing process, which involves refining prose and rectifying grammatical and lexical errors. Additionally, LLMs are utilized for identifying pertinent related works. All content produced by these models undergoes human verification prior to its inclusion in the manuscript.

\end{document}